# Graphes paramétrés et outils de lexicalisation

Éric Laporte, Sébastien Paumier

IGM – Université de Marne-la-Vallée, 5 bd Descartes
77454 Marne-la-Vallée CEDEX 2
eric.laporte@univ-mlv.fr, sebastien.paumier@univ-mlv.fr

**Résumé**  La lexicalisation des grammaires réduit le nombre des erreurs d'analyse syntaxique et améliore les résultats des applications. Cependant, cette modification affecte un système d'analyse syntaxique dans tous ses aspects. Un de nos objectifs de recherche est de mettre au point un modèle réaliste pour la lexicalisation des grammaires. Nous avons réalisé des expériences en ce sens avec une grammaire très simple par son contenu et son formalisme, et un lexique syntaxique très informatif, le lexique-grammaire du français élaboré au LADL. La méthode de lexicalisation est celle des graphes paramétrés. Nos résultats tendent à montrer que la plupart des informations contenues dans le lexique-grammaire peuvent être transférées dans une grammaire et exploitées avec succès dans l'analyse syntaxique de phrases.

**Abstract**  Shifting to a lexicalized grammar reduces the number of parsing errors and improves application results. However, such an operation affects a syntactic parser in all its aspects. One of our research objectives is to design a realistic model for grammar lexicalization. We carried out experiments for which we used a grammar with a very simple content and formalism, and a very informative syntactic lexicon, the lexicon-grammar of French elaborated by the LADL. Lexicalization was performed by applying the parameterized-graph approach. Our results tend to show that most information in the lexicon-grammar can be transferred into a grammar and exploited successfully for the syntactic parsing of sentences.

**Mots-clés :**  Grammaires, syntaxe, lexicalisation, réseaux de transitions récursifs
**Keywords:**  Grammar, syntax, lexicalization, recursive transition networks

# 1 Introduction

L'analyse syntaxique automatique du texte en langue naturelle nécessite l'utilisation de grammaires syntaxiques et sémantiques formalisées. La majorité des auteurs de grammaires a longtemps fait porter l'essentiel de ses efforts sur la généralité des règles qu'ils y formalisaient. Depuis une quinzaine d'années, beaucoup tentent au contraire de prendre en compte les interdépendances massives entre éléments lexicaux et règles grammaticales, et de lexicaliser les grammaires, c'est-à-dire d'y introduire des informations spécifiques à certains éléments lexicaux. Cela réduit le nombre d'erreurs d'analyse syntaxique (Briscoe, Carroll, 1993 ; Carroll, Fang, 2004) et améliore les résultats des systèmes de traduction automatique (Han et al., 2000) ou de réponse à des questions (Jijkoun et al., 2004). Cependant, cette modification affecte un système d'analyse syntaxique dans tous ses aspects : lexique, structures de données, algorithmes et ordres de grandeur des objets. Un de nos objectifs de recherche est de mettre au point un modèle réaliste pour la lexicalisation des grammaires syntaxiques et sémantiques. Nous avons réalisé des expériences en ce sens avec une grammaire très simple par son contenu et son formalisme, et un lexique syntaxique très informatif, le lexique-grammaire du français élaboré au LADL[1] (Gross, 1984, 1994). La méthode de lexicalisation est celle des graphes paramétrés. Dans cette présentation, après un rappel sur des travaux antérieurs, nous décrivons notre formalisme grammatical, la lexicalisation par graphes paramétrés et nos expériences d'évaluation, et nous en tirons des conclusions[2].

# 2 Travaux antérieurs

L'approche la plus répandue pour la lexicalisation consiste à enrichir une grammaire générique[3] par apprentissage automatique à partir d'exemples trouvés dans un corpus arboré (Neumann, 1998). Le principal avantage de ce principe est sans doute la division du travail entre deux types de tâches : d'une part, la construction des corpus arborés, qui demande essentiellement des talents en linguistique, et d'autre part l'apprentissage automatique, envisagé comme un problème presque purement informatique. La difficulté de l'entreprise ressort, cependant, si l'on met en relation la complexité des comportements à apprendre et le nombre limité d'exemples présents dans les corpus arborés disponibles. La couverture lexicale et la couverture grammaticale d'un corpus arboré sont en effet toutes deux tributaires de sa taille[4]. De plus, la séparation d'un mot ambigu en plusieurs entrées lexicales, notion fondamentale, est peu développée dans les corpus arborés disponibles.

Une deuxième approche (Egedi, Martin, 1994 ; Abeillé, Candito, 2000) consiste à enrichir manuellement des grammaires génériques en informations relatives à des éléments lexicaux.

---

[3] Dans la suite, nous appelons grammaires génériques les grammaires non lexicalisées.

[4] La cible de l'apprentissage étant la correspondance entre éléments lexicaux et constructions grammaticales, le nombre d'exemples pertinent ici est plutôt une couverture lexico-grammaticale qui nécessiterait des quantités de corpus arborés d'un ordre de grandeur encore supérieur.

Le fait que la procédure soit manuelle permet aux auteurs des ressources d'exercer un contrôle direct sur leur contenu, aussi bien lors de la maintenance et de la correction d'erreurs que lors de la construction initiale. De tels projets visent à la construction de ressources de qualité et nécessitent de nombreuses années de travail par des équipes qui combinent des compétences réelles en informatique et en linguistique.

Le travail présenté ici se situe dans une troisième approche (Roche, 1993, 1999). Il s'agit d'introduire dans une grammaire des informations présentes dans un lexique syntaxique et sémantique existant, construit manuellement. Le principal avantage de cette méthode vient de l'utilisation d'un lexique. La construction manuelle de lexiques se prête bien à la prise en compte de deux notions cruciales : d'une part, les unités lexicales multi-mots[5] ; d'autre part, la séparation d'un mot ambigu en plusieurs entrées lexicales. Comme la précédente, cette approche comporte une exigence de qualité et suppose une combinaison de compétences en informatique et en linguistique. Elle nécessite, pour aboutir, que le lexique ait une couverture lexico-grammaticale plus étendue qu'aucun lexique disponible à l'heure actuelle. Cependant, les enjeux sont à la hauteur de la difficulté, et le lexique-grammaire du LADL est une approximation pertinente du lexique requis. (Silberztein, 1999) a modifié l'approche de Roche par l'utilisation systématique de réseaux de transitions récursifs (RTN), l'introduction d'outils de visualisation et d'édition de RTN, la simplification des outils de traitement des grammaires, et l'adoption d'un autre algorithme d'analyse syntaxique. Nous avons poursuivi cette recherche avec de nouvelles grammaires et réalisé des expériences d'évaluation. D'autres projets du même type sont en cours (Gardent et al., 2005).

# 3   Les RTN comme formalisme grammatical

Les RTN ont un statut à part parmi les formalismes de grammaires syntaxiques et sémantiques en raison de leur simplicité biblique. Il s'agit d'un outil purement formel, dépourvu de notions linguistiques, au même titre que les automates finis ou les grammaires algébriques. Le graphe de la fig. 1, par exemple, ne comporte explicitement ni traits, ni syntagmes, ni règles d'unification, ni arbres syntaxiques. C'est le composant principal d'une grammaire générique des phrases déclaratives transitives directes à verbe distributionnel[6] et à un seul complément essentiel. Les arêtes du graphe représentent la concaténation de formes linguistiques. Les listes contenues dans les noeuds, comme *que/qu'*, représentent des paradigmes. Les identificateurs affichés sur fond gris, comme *N0*, représentent des appels à d'autres graphes. Les textes apparaissant hors des noeuds, comme *Luc a vu le lit*, sont des commentaires destinés à faciliter la lecture. Les formes contenues dans les noeuds sont ici des mots mais peuvent aussi être des masques lexicaux (Blanc, Dister, 2004), et donc comporter des traits. La quasi-absence de métalangage (uniquement les noms des graphes et les traits) est un facteur important de lisibilité. Cette forme graphique de RTN, inspirée du système Intex (Silberztein, 1993), se prête à la construction manuelle de grammaires. En particulier, la structuration en graphes petits et lisibles[7] facilite l'accumulation de composants peu dépendants les uns des autres.

---

[5]   Ce modèle a été utilisé pour l'analyse syntaxique de phrases à prédicats multi-mots (Senellart, 1999).

[6]   Un verbe distributionnel, par exemple *préférer* dans *Luc préfère le thé au café*, est un verbe qui n'est ni un verbe support, comme *avoir* dans *Luc a une préférence pour le thé*, ni figé avec un de ses actants, comme *rendre* dans *Luc se rend compte de la difficulté*. Sur ces notions, v. Gross (1981) et Leclère (2002).

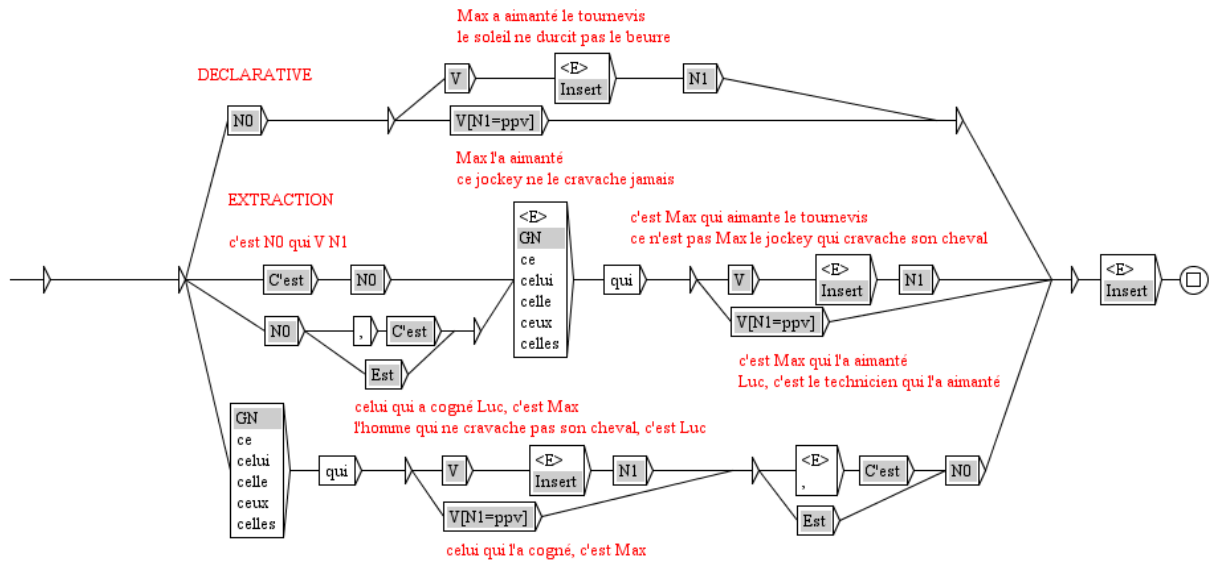

Figure 1 : Graphe générique

## 4   Lexicalisation par graphes paramétrés

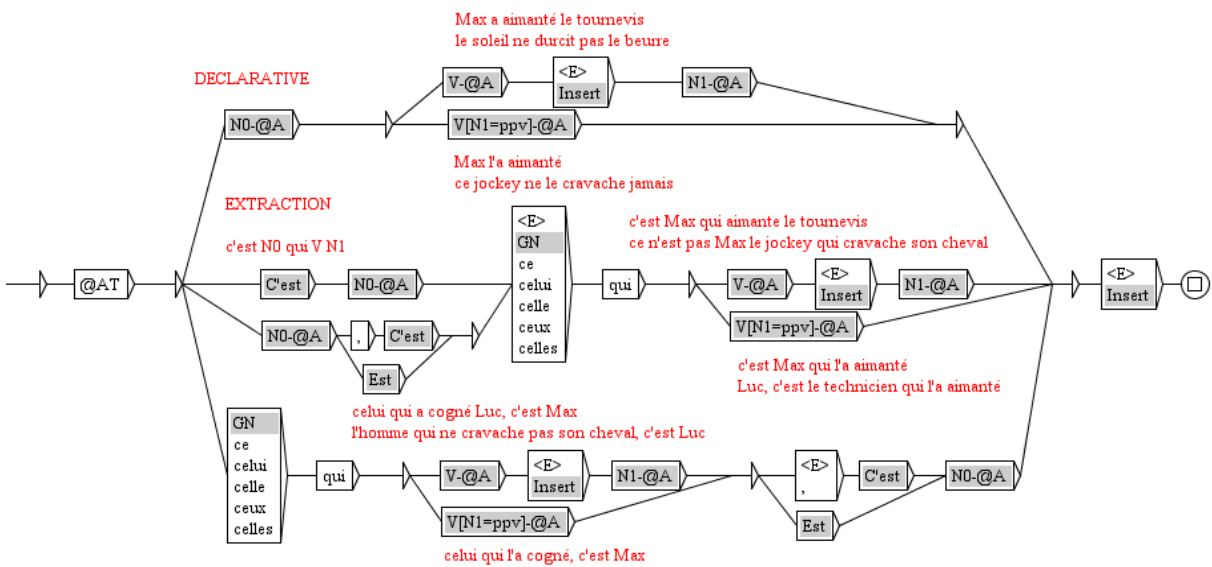

Figure 2 : Graphe paramétré

Le graphe de la figure 2 est une variante du précédent, paramétrée en vue de sa lexicalisation. La seule différence est la présence des paramètres @AT, près de l'état initial, et @%, dans certains identificateurs de sous-graphes. La figure 3 est le résultat de la lexicalisation pour le verbe *préférer* : le paramètre @AT a été remplacé par un noeud intermédiaire vide, et les paramètres @% par le numéro identificateur de ce verbe. Le principe de la lexicalisation par graphes paramétrés est ainsi la génération de variantes lexicalisées d'un graphe générique. On insère manuellement des paramètres dans les graphes de la grammaire générique (Paumier, 2001). Les variantes lexicalisées sont engendrées automatiquement par le système Unitex[8] (Paumier, 2003) à partir du graphe paramétré. Lors de cette opération les paramètres sont remplacés par des valeurs trouvées dans une table du lexique-grammaire[9]. Les graphes lexicalisés obtenus sont les adaptations du graphe générique aux différentes valeurs que peut prendre un mot. Celui-ci est appelé l'*ancre* du graphe paramétré. Chaque valeur de l'ancre fait l'objet d'une entrée de la table. Dans notre exemple, l'ancre est le verbe principal de la phrase. Unitex produit également un graphe qui appelle toutes les variantes lexicalisées du graphe générique. Ces opérations sont indépendantes du texte à analyser et produisent typiquement des centaines de milliers de graphes. Pour faciliter l'exploitation de ce type de grammaire, nous étudions plusieurs solutions, notamment l'utilisation d'algorithmes parallèles et l'adaptation de la grammaire au texte.

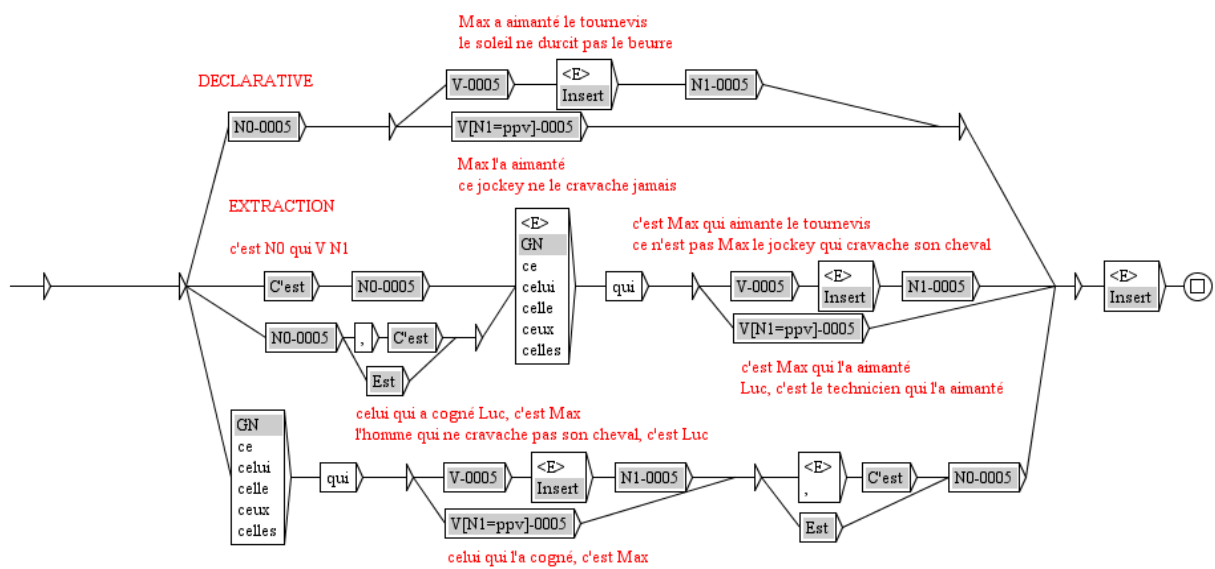

Figure 3 : Graphe lexicalisé pour le verbe *préférer*

# 5   Types de paramètres

Les paramètres de lexicalisation signalent les éléments d'un graphe qui varient en fonction de l'ancre : valeurs lexicales, interrupteurs de chemins... En voici une typologie.

## 5.1   Valeurs lexicales

Certaines valeurs lexicales dépendent de l'ancre : par exemple, des prépositions régies par des prédicats (*préférer à, retenir de*), ou encore l'ancre elle-même. Ainsi, dans la figure 4, le paramètre @D correspond au lemme de l'ancre. Remplacé dans l'expression <@D.*V:K*>, il

---

produit le masque lexical *<préférer.V:K>* qui code les formes du participe passé (*:K*) du verbe (*.V*) *préférer*[10].

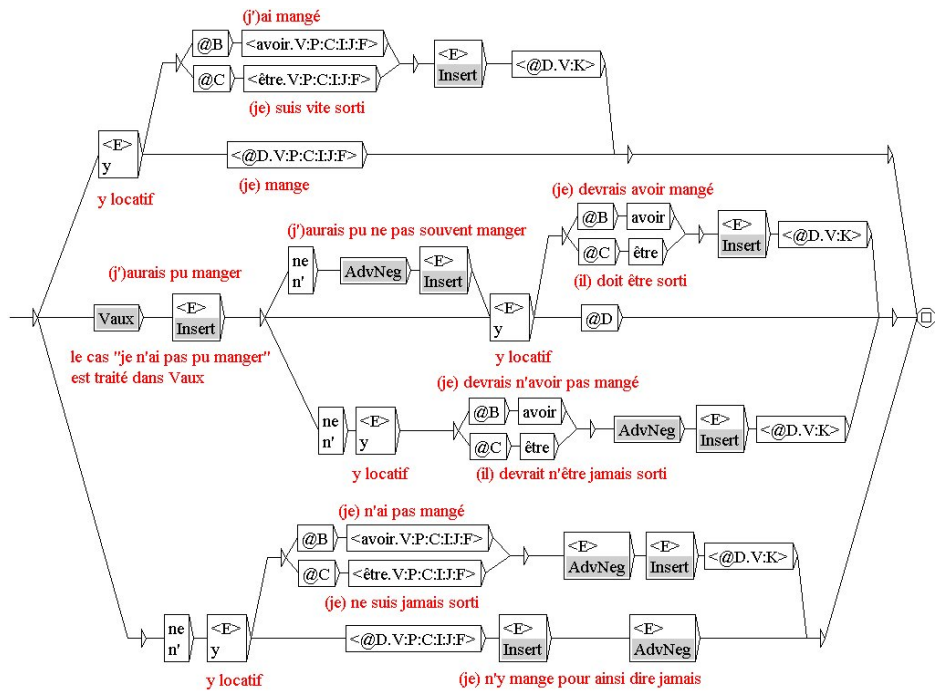

Figure 4 : Le graphe paramétré Vpar

## 5.2 Interrupteurs de chemins

L'acceptabilité de certaines constructions dépend de l'ancre. Cette dépendance est codée par un interrupteur de chemin, tel que le paramètre @AT dans la fig. 2. La table indique par un trait binaire pour chaque entrée si cette construction est acceptable. Suivant le cas, le paramètre est remplacé soit par un noeud intermédiaire vide (fig. 3), qui assure la connexion entre le début et la fin du chemin, soit par une valeur nulle, qui l'interrompt[11]. Les paramètres @B et @C dans la figure 4 sont également des interrupteurs de chemins.

## 5.3 Identificateurs d'entrées et de composants de la grammaire

Lorsque plusieurs graphes d'une même grammaire sont lexicalisés en fonction d'une même ancre, comme le graphe principal et les graphes N0, V, V[N1=ppv] et N1 de la fig. 1, la combinaison des différents graphes lexicalisés nécessite que leurs noms contiennent des identificateurs des entrées de la table. Dans notre exemple, il s'agit de 0005, produit par le paramètre @%.

---

[10]  Un graphe paramétré peut aussi ne pas contenir l'ancre. Ainsi, dans la fig. 3, l'ancre est le verbe principal ; le sous-graphe N0-0005 représente son sujet et tient compte des contraintes exercées sur le sujet par une des valeurs lexicales du verbe, mais le verbe ne figure pas dans ce sous-graphe.

[11]  Pour améliorer la lisibilité des graphes paramétrés, les notations peu parlantes @AT, @D seront remplacées par les intitulés des propriétés : respectivement, @N0 V N1@ et @V@.

Dans la fig. 3, le graphe lexicalisé appelle des sous-graphes adaptés à l'ancre, en raison des contraintes de sélection que celle-ci exerce sur le sujet. Cet effet est obtenu en insérant un paramètre de lexicalisation dans le nom du graphe appelé (fig. 2). Le sous-graphe appelé doit être lexicalisé par ailleurs.

## 5.4 Niveaux de généricité dans la lexicalisation

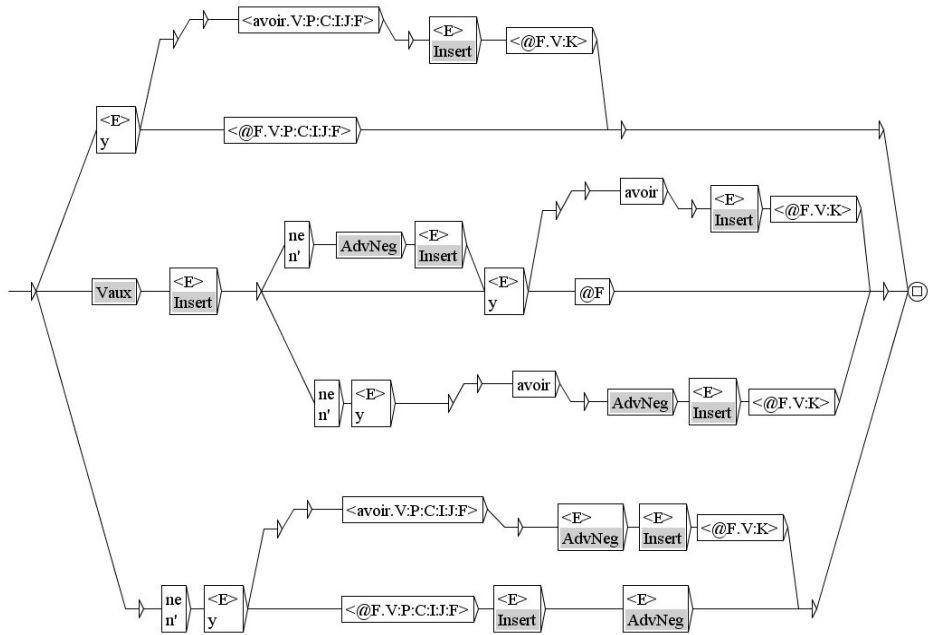

Figure 5 : Le graphe Vpar lexicalisé pour la classe 4

Dans les exemples précédents, la lexicalisation d'un composant d'une grammaire générique produit des variantes adaptées à des éléments lexicaux spécifiques. Il est parfois préférable qu'elle produise des variantes adaptées à des classes d'éléments lexicaux. C'est en fait le cas du graphe de la fig. 4 : il est possible de classer les verbes en sous-catégories telles que certains de ses paramètres dépendent assez souvent de l'appartenance du verbe aux sous-catégories, et non de sa valeur lexicale précise. C'est ce qui a été fait lors de la construction du lexique-grammaire du français. On peut donc utiliser ce graphe pour une lexicalisation par classes. Lors de la génération des variantes lexicalisées des graphes, on recherche les valeurs des paramètres de lexicalisation dans une table comportant une entrée par classe[12]. Par exemple, tous les verbes de la table 4 (Gross, 1975) sélectionnent obligatoirement l'auxiliaire *avoir* pour les temps composés. Le résultat de la lexicalisation pour cette classe (fig. 5) comporte donc des valeurs fixes à la place des paramètres @B et @C de la fig. 4. En revanche, dans ce même graphe, la valeur lexicale du verbe n'est pas une valeur fixe. Elle apparaît donc sous la forme d'un paramètre (@F) pour une passe de lexicalisation ultérieure.

---

[12]    En l'occurrence, il s'agit de la table générale des verbes distributionnels.

# 6   Evaluation

Une évaluation précise de cette méthode de lexicalisation ne pourra être effectuée qu'en présence de plusieurs ressources : une grammaire d'une couverture raisonnable ; les tables de lexique-grammaire décrivant les classes d'éléments lexicaux ayant le statut d'ancres ; les tables générales décrivant les propriétés des tables précédentes ; les lexiques donnant pour les éléments lexicaux les valeurs des traits syntaxiques et sémantiques utilisés dans la grammaire. Aucune de ces ressources n'étant disponible en totalité, nous avons effectué une expérience avec des ressources partielles : un jeu de 20 graphes pour les phrases déclaratives et de 54 graphes qui décrivent grossièrement les groupes nominaux, les verbes modaux et les insertions adverbiales ; la table 32NM du lexique-grammaire (Boons et al., 1976) ; l'entrée de cette dernière dans la table générale des verbes distributionnels ; et les lexiques de formes fléchies du LADL. Nous avons traité *Le Tour du monde en 80 jours* de Jules Verne (72 000 mots) avec Unitex. La grammaire générique retient 2014 occurrences, dont 3,8 % sont des identifications correctes[13] d'occurrences de la table 32NM. En lexicalisant les 20 graphes pour les phrases déclaratives, on retient seulement 486 occurrences, ce qui fait passer la précision à 16 %, soit 4 fois plus. Le bruit résiduel se répartit en :

- 34 % dû aux approximations de la grammaire ;

- 27 % dû à diverses ambiguïtés lexicales (expressions figées, *il y a*, *A* pour *À*...)[14] ;

- 24 % dû à des ambiguïtés des verbes de la table 32NM avec d'autres entrées verbales.

Il serait nécessaire de formaliser plus de contraintes dans la grammaire pour augmenter encore la précision.

Pour évaluer le rappel, nous avons fait une seconde expérience avec une grammaire plus réduite et une seule entrée lexicale de verbe. Nous avons lexicalisé un graphe légèrement plus élaboré que celui de la fig. 2 pour l'entrée du verbe *receler* décrite dans la table 32NM et appliqué la grammaire obtenue à un texte journalistique de 20 millions de mots. Du fait que les formes conjuguées de ce verbe ne sont pas ambiguës avec d'autres catégories grammaticales, nous obtenons 98 % de précision : sur les 50 constructions reconnues, on compte 1 erreur due à une forme interrogative avec inversion du sujet. Notre corpus contenait 137 occurrences de ce verbe, dont 47 correspondant à des participes présents, à des infinitives et à des relatives qui ne sont pas prévus par la grammaire. Sur les 90 autres occurrences, notre grammaire en reconnaît correctement 54 %, et en ignore :

- 21 % à cause d'insertions non prévues, de parenthèses ou de guillemets :
  *ce " musée à l'intérieur du musée " recèle désormais quelque treize cents oeuvres*

- 24 % à cause d'autres lacunes de la grammaire (graphe de groupe nominal incomplet, coordinations de phrases, etc.) :
  *il ne recèle rien de médiocre*

---

[13] Parmi celles-ci, 62 % reconnaissent complètement la structure de la phrase et 38 % partiellement.

[14] Ces valeurs ne sont pas comparables à ceux fournis par d'autres analyseurs syntaxiques, car notre étiquetage morpho-syntaxique a plus de rappel mais moins de précision que ceux utilisés par les autres auteurs.

> *Il voyage beaucoup <u>mais</u> <u>recèle</u> des tendances casanières*
> *Ses gravures ne <u>recèlent</u> cependant <u>ni citation, ni imitation</u>*

Le silence est donc dû à la grammaire générique, et non à la méthode de lexicalisation. Voici quelques exemples d'occurrences correctement analysées :

> *Constantine est une ville qui <u>recèle</u> des trésors culturels*
> *Les bordereaux <u>recèlent</u> enfin les noms de plusieurs grandes entreprises françaises*
> *Les instruments financiers dérivés en général <u>recèlent</u> des risques importants, parfois*
> *masqués derrière leur appareil technique et mathématique*
> *La banlieue parisienne <u>recèle</u> bien des surprises en matière de handball*

# 7   Conclusion

Les résultats obtenus tendent à confirmer les résultats précédents sur le réalisme du modèle de lexicalisation par graphes paramétrés : la plupart des informations contenues dans le lexique-grammaire peuvent être transférées dans une grammaire générique, et exploitées avec succès dans l'analyse syntaxique de phrases.

Par ailleurs, ce modèle de lexicalisation est fortement découplé du formalisme grammatical. Les notions auxquelles il fait appel : valeurs lexicales, interrupteurs de chemins, identificateurs d'entrées lexicales ou de composants de la grammaire, existent dans tout formalisme grammatical ou peuvent facilement y être incorporés. Les potentialités de ce modèle nous semblent donc beaucoup plus étendues que l'utilisation qui en a été faite depuis sa création. En particulier, des expériences sont en cours dans notre équipe (Blanc, Constant, 2005) sur l'application de ce modèle à un formalisme grammatical plus complexe, avec unification et production d'un arbre distinct de l'arbre de dérivation, et avec l'algorithme d'analyse syntaxique d'Earley, plus efficace.

# Références